\documentclass{article}
\usepackage{amssymb}
\usepackage{times}

\usepackage[final]{corl_2025} 
\usepackage{multicol}
\usepackage{wrapfig}
\usepackage{amsmath,amssymb,amsfonts}
\usepackage{algorithmic}
\usepackage{graphicx}
\usepackage{textcomp}
\usepackage{xcolor}

\usepackage{fancyhdr}

\usepackage{subcaption}
\usepackage{stfloats}
\usepackage{booktabs}
\usepackage{makecell}
\usepackage{multirow}
\usepackage{color}
\usepackage[algo2e, ruled]{algorithm2e}
\usepackage{todonotes}

\hypersetup{hidelinks,bookmarks=true}

\newcommand{\algo}{{\sc\textsf{HBC}}}

\title{Bipedal Balance Control with Whole-body Musculoskeletal Standing and Falling Simulations}

\author{
Chengtian Ma, Yunyue Wei, Chenhui Zuo, Chen Zhang, Yanan Sui \\
\\
Tsinghua University
}

\begin{document}
\maketitle
\
\begin{abstract}

Balance control is important for human and bipedal robotic systems. While dynamic balance during locomotion has received considerable attention, quantitative understanding of static balance and falling remains limited. This work presents a hierarchical control pipeline for simulating human balance via a comprehensive whole-body musculoskeletal system. We identified spatiotemporal dynamics of balancing during stable standing, revealed the impact of muscle injury on balancing behavior, and generated fall contact patterns that aligned with clinical data. Furthermore, our simulated hip exoskeleton assistance demonstrated improvement in balance maintenance and reduced muscle effort under perturbation. This work offers unique muscle-level insights into human balance dynamics that are challenging to capture experimentally. It could provide a foundation for developing targeted interventions for individuals with balance impairments and support the advancement of humanoid robotic systems. Project page: \url{https://lnsgroup.cc/research/bipedal_balance}.
\end{abstract}

\keywords{balance control, bipedal standing and falling, musculoskeletal system}

\section{Introduction}
\label{sec:introduction}

Bipedal locomotion and balance have been extensively studied in robotics and control, with research demonstrating the inherent challenges of stabilizing underactuated, high-degree-of-freedom systems \cite{fujimoto1998robust, sugihara2009standing}. Robotic bipedal stability during locomotion has been analyzed through criteria like limit cycles and gait periodicity \cite{gubina1974dynamic, pratt1998intuitive}. While bipedal standing seems naturally achievable, stable standing remains a challenging problem, particularly when accounting for double-support phases and area contact \cite{contact2018mummolo}. However, the foundational behavior enabling bipedalism, human static standing, remains understudied due to the lack of models capable of capturing its musculoskeletal complexity. Humans rely on an obligate bipedal stance, requiring precise coordination between neuromuscular control and biomechanical dynamics. Unlike occasional bipeds seen in other animals, human standing is characterized by a vertical spine, the absence of auxiliary balancing structures (e.g., prehensile tails), and an upper body for multitasking while maintaining balance \cite{SKOYLES20061060}. This skill is acquired over years of development, relying on advanced neural integration and reflexes \cite{diener1983reflex}. Mastering static balance could lag behind walking in children \cite{breniere1998development}, highlighting its difficulty. The process integrates multisensory inputs and fine-grained muscle coordination \cite{massion1998postural}, posing a control challenge distinct from robot locomotion.


 Humans achieve remarkable balance adaptability, but the principles underlying this robustness remain poorly understood. The human musculoskeletal system, with its high degrees of freedom and nonlinearity, requires more sophisticated control mechanisms to perform the dynamical control of stable standing. A critical gap persists: \textit{no prior work has systematically investigated bipedal standing control using a full-body musculoskeletal model}, which is essential to decode fine-grained, muscle-level postural dynamics \cite{valero2009computational}. 

In contrast to maintaining balance, falling is a leading cause of injury among aging populations \cite{kakara2024cause}, resulting in both psychological trauma and physical harm \cite{luque2014comparison}. Despite its significance, falling dynamics remain poorly understood due to data scarcity and ethical constraints in real-world experiments \cite{CHOI2015911}.

In this work, we introduce a hierarchical control and analysis method for full-body bipedal balance control. Our framework, deployed on a full-body human musculoskeletal model with 700 muscle-tendon units~\cite{tsht}, enables biomechanically plausible simulations of standing balance and falls that would be limited in real experiments. Through extensive simulations, we identify specific dynamical balance behavior during stable standing, reveal the adaptive balancing strategies under injury conditions, and demonstrate how hip exoskeleton assistance can enhance balance while reducing muscular effort. We provide muscle-level insights during balance that remain inaccessible through conventional experimental methods. To the best of our knowledge, our approach is the first to leverage a high-dimensional, whole-body musculoskeletal model to systematically study human balance dynamics at muscle-level resolution, providing insights for both biomechanical understanding of bipedal balance and assistive technology development.

\begin{figure*}[h]
  \centering
  \includegraphics[width=1\linewidth]{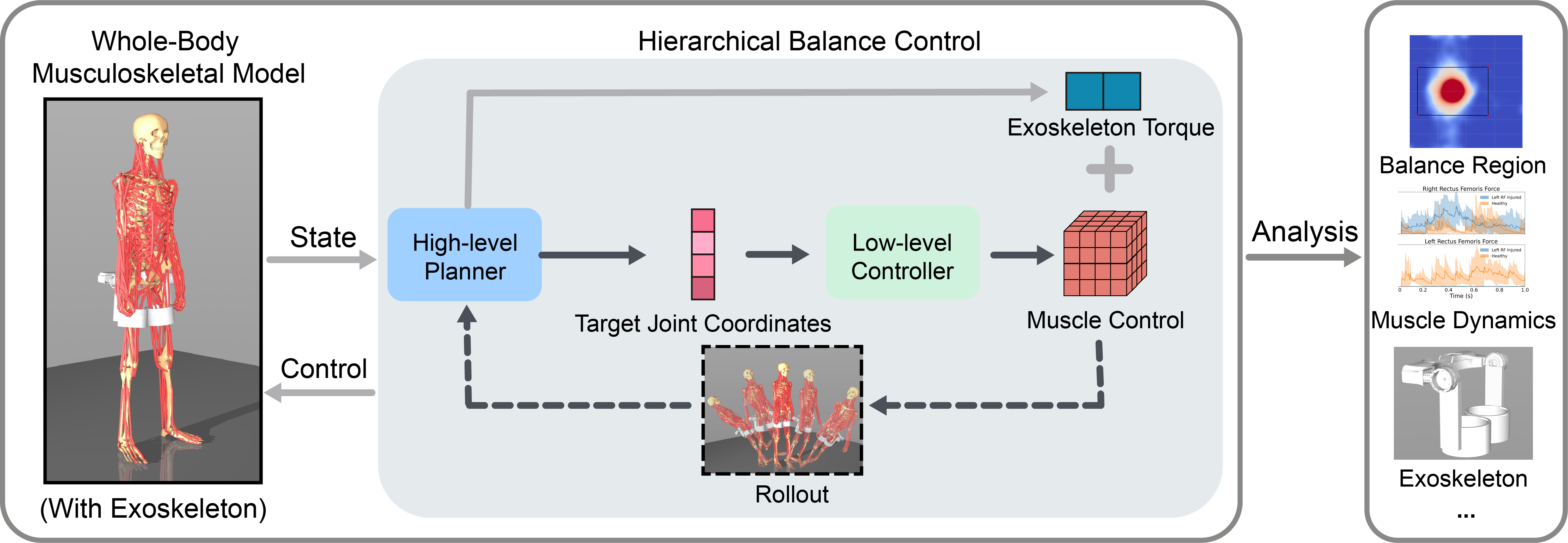}
  \caption{Control and analysis of human musculoskeletal standing and falling. \algo~enables training-free balance control and efficient collection of dynamical behavior for balance analysis. Our method supports concurrent control planning for exoskeleton-assisted scenarios, facilitating integrated evaluation and optimization of human-exoskeleton interaction.}
  \label{fig:standing}
\end{figure*}

\section{Related Work}
\label{sec:relatedwork}

\subsection{Bipedal Balance from Robotics to Human Musculoskeletal Models}

The balance control of bipedal robots has been extensively studied, with numerous approaches demonstrating the inherent challenges of stabilizing underactuated systems. Traditional research has employed simplified models such as single inverted pendulum \cite{morasso2019quiet} and double inverted pendulum models \cite{SUZUKI201255, gunther2011watching} to represent bipedal balance dynamics by focusing on the Center of Mass (CoM) motion and the Zero Moment Point (ZMP). 


However, human balance control is far more complex than robot models can capture, involving intricate sensorimotor integration executed through high-dimensional muscular systems \cite{horak2006postural} rather than simplified bipedal standing frameworks like ZMP strategies \cite{stabilityregion2017kim, vukobratovic2004zero}. A gap between theoretical models and biological reality remains, necessitating the use of musculoskeletal models \cite{delp2007opensim, MyoSuite2022}. Recent work introduced a full-body human musculoskeletal system that simulates whole-body dynamics \cite{tsht}. Control methods using deep reinforcement learning (DRL) have been explored for these high-dimensional systems \cite{2019mass, deprl, feng2023musclevae, berg2024sar, he2024dynsyn}. While progress has been made in improving locomotion capabilities, the control of full-body standing balance remains unexplored.

\subsection{Investigation of Human Balance and Fall}

 While robotic fall-prediction systems have advanced through multi-sensor fusion \cite{subburaman2019human} and machine learning \cite{mungai2024fall}, human fall dynamics lack physiologically accurate understanding. Existing datasets \cite{HSIAO19971, CHOI2015911} primarily consist of voluntary falls, which differ substantially from real-world falls in terms of impact dynamics and protective responses \cite{KLENK2011368}. 

Exoskeletons have been used to assist balance for various populations \cite{farkhatdinov2019assisting, 8488066}, but experimental studies still have significant limitations. Most exoskeleton research has focused on locomotion tasks \cite{luo2024experiment, lu2025human}, with limited investigation into balance scenarios. Studies that target at assisting balance with exoskeletons typically involve small participant numbers and controlled lab conditions \cite{8488066, monaco2017ecologically}. These limitations in experimental scope and participant diversity hinder a comprehensive understanding of how exoskeletons might benefit the balance control.
Musculoskeletal simulations have aided exoskeleton design \cite{molinaro2024task} by estimating joint torques \cite{lam2021joint} or serving as control testbeds \cite{mo2021simulation}. To the best of our knowledge, there are no existing works that use musculoskeletal simulation to test and validate the performance of balance-targeted exoskeletons. 

\section{Problem Setting}
\label{sec:problemsetting}

\subsection{Musculoskeletal Model Dynamics} 
\label{sec:msmodeldynamics}
We use the full-body \href{https://lnsgroup.cc/research/MS-Human}{\textcolor{blue}{MS-Human-700}} model \cite{tsht}, which comprises of 90 rigid body segments, 206 joints and 700 muscle-tendon units. It is implemented in the MuJoCo physics engine \cite{mujoco}. The actuators of the model are 700 Hill-type \cite{zajac1989muscle} muscles. The force generated by each actuator, and the temporal relation between muscle activation $act$ and the input control signal $u$ can be described by the following equations:
\begin{align}
f_m(act)&=f_{max}\cdot [F_{l}(l_m)\cdot F_v(v_m)\cdot act + F_p(l_m)], \quad
    \frac{\partial act}{\partial t} = \frac{u-act}{\tau(u,act)},\label{equ:muscle_dynamic}
\end{align}
where $F_l$ and $F_v$ are active force-length and force-velocity functions, $F_p$ is the passive force-length function, and $l_m, v_m$ are normalized muscle length and velocity. $f_{max}$ is the maximum isometric muscle force. Muscle activation $act$ is calculated with control $u$ in Eq. (\ref{equ:muscle_dynamic}) \cite{millard2013flexing}.

\subsection{Standing Task Design and Balance-Related Scenarios}

We treat the human standing control problem as a finite horizon Markov decision process with state $s\in \mathcal{S}$, control $u\in \mathcal{U}$, dynamics $s_{t+1} = f(s_t, u_t)$, and time step $t$. For a given initial state $s_0$ and a desired standing horizon $T$, we aim to find a control sequence $U^{\star}_T=(u_0, ..., u_{T-1})$ that successfully maintains the standing posture by minimizing a pre-defined cost function $C$:
\begin{align}
    \label{eq: problem}
    U^{\star}_T = \text{argmin}_{U_T}\sum_{t=0}^{T-1} C(s_t, u_t)
\end{align}
The control goal for standing is: (1) Near-zero CoM horizontal velocity and (2) CoM within the support region, ensuring no flipping tendency. The model may step for balance, but its initial position remains constant across experiments.

\section{The Control Algorithm and Analysis for Static Balancing}
\label{sec:method}

In this section, we present our control and analysis method for whole-body static balance in humans. We begin by introducing a hierarchical control framework that enables training-free control of high-dimensional balance dynamics. We then detail our analysis for evaluating balance performance under varying physiological conditions and with assistance.

\begin{algorithm2e}[t]
\SetAlgoLined
\DontPrintSemicolon
\LinesNumbered
\caption{
Hierarchical Balance Control (\algo)
}

\label{alg:hbc}

\KwIn{Model dynamics $f$, total time steps $T$, execution length $t_e$, rollout horizon $h$, particle number $n$, MPPI iteration number $r$, initial distribution parameter $\mu, \sigma$}
\KwOut{Action sequence $U^*$}

$s_0\sim \mathcal{S}_0, U^*\leftarrow \emptyset$\;
\For {$t = 0, \cdots, T-1$}{
    \If{$t \bmod t_e = 0$}{
        \For {$i = 1, \cdots, r$}{
         $z_1, ..., z_N \sim \mathcal{N}(\mu,\sigma)$\;
            $c_1, \cdots, c_n \leftarrow \text{Rollout}(z_1, \cdots, z_n)$\;
        Update $\mu, \sigma$ using e.q. (\ref{eq: mu})\;
        }
        $z^*\sim \mathcal{N}(\mu,\sigma)$\; 
    }
    $u_{t}\leftarrow \pi(s_{t}, z^*)$\;
    $U^* = U^* \cup u_t$\;
    $s_{t+1}\leftarrow f(s_{t}, u_{t})$
    
}
\end{algorithm2e}





\subsection{Hierarchical Balance Control Algorithm}
\label{sec:hbc}

We propose Hierarchical Balance Control (\algo), a hierarchical algorithm for the balance control of human body. As shown in Figure \ref{fig:standing}, a high-level planner first proposes a set of major joint coordinates $z^* \in \mathcal{Z}$ as the target of the low-level controller, $\pi(u|s, z)$, which coordinates muscle controls to achieve the target joint coordinates. \algo~uses model predictive control (MPC) to optimize the target joint coordinates according to the cumulative cost function over a short horizon: 
\begin{align}
    \label{eq: mpc}
    z^{\star} = \text{argmin}_{z}\sum_{t=t_p}^{t_p+H-1} C(s_t, u_t), u_t = \pi (s_t, z),
\end{align}
where $t_p=\{t \mid t \mod t_e = 0\}$ is the planning timestep and $H \ll T$ is the planning horizon. Compared to vanilla MPC methods, \algo~reduces the parameter space from the scale of muscle number to joint number, which is a feasible dimension for effective planning and control. 

For the high-level controller, we integrate Model Predictive Path Integral (MPPI, \cite{mppi2016}) as a sampling-based planner. At the beginning of each rollout process, $N$ sets of target joint angles $z_1, z_2, ..., z_N$ are sampled from the current target distribution $\mathcal{N}(\mu,\sigma)$ to generate $N$ trajectories.  The cumulative cost $c_n = \sum_{t=t_p}^{t_p+H-1} C(s_t, \pi(s_t, z_n))$ is calculated from the parallel rollout trajectory of target $z_n$. The new mean and covariance are calculated based on the weighted average of top $k$ targets $(z_1^{*}, z_2^{*}, ..., z_k^{*})$ whose corresponding trajectories return the minimal costs. The update is implemented as follows:
    \begin{equation}
    \label{eq: mu}
         \mu = \frac{\sum_{j=1}^{k}w_j \cdot z_j^{*} }{\sum_{j=1}^{k}w_j}, \sigma = \sqrt{\frac{\sum_{j=1}^{k}w_j \cdot (z_j^{*}-\mu)^2 }{\sum_{j=1}^{k}w_j}},
    \end{equation}
where $w_j=e^{-\frac{1}{\lambda}}c_j$ is the aggregate weight for $z_j$, and $\lambda$ is the softmax temperature. The full method of getting control sequences is shown in Algorithm \ref{alg:hbc}.


For the low-level controller, we use a normalized PD control-like formulation over muscle lengths to derive desired muscle forces \cite{feng2023musclevae}:
\begin{equation}
        f_m = min(0,k_p\cdot(l^{*}_m-l_m)/l_{range} + k_d\cdot(0 - \Dot{l}_m)),
    \end{equation}
where $f_m$ stands for muscle forces, $k_p$ and $k_d$ are PD control gains, $l_m$ stands for actual muscle lengths, $\Dot{l}_m$ stands for muscle velocities and $l^{*}_m$ is target muscle lengths. The difference between the maximum muscle length and the minimum muscle length of each muscle, $l_{range}$, is used to normalize muscle lengths and stabilize control effect. Given desired muscle forces, the desired muscle control can be derived according to the muscle actuator dynamics in e.q. (\ref{equ:muscle_dynamic}) to facilitate full-body control.

Compared to DRL-based methods, \algo~is a training-free method which generates effective controls in minutes. This advantage enables the collection of large dynamical dataset during static balance control under different conditions for statistical analysis.

\subsection{Standing and Fall under Different Physiological Conditions}
\label{sec:injury}

We apply \algo~over the MS-Human-700 model to collect dynamical balance behavior during stand control. The model was initialized in a natural standing posture and then stood for 5 seconds. We consider the following criteria for balance and fall monitoring:


\textbf{Balance behavior monitoring.} During each simulation, the model is considered to have achieved stable standing if two conditions were met: (1) No body parts other than the feet contacted the ground. (2) The CoM didn't leave the real-time support polygon when 5s is reached. We collected CoM positions, muscle force of the model and body posture data at a 500 Hz frequency for analysis.

\textbf{Fall detection and recording.} When we apply \algo~from an up-straight posture for 5 seconds, the human model has small possibility to fall from standing. We track the CoM position of the MS-Human-700 model and the support polygon formed by foot contact points (heels and toes). We define fall duration as the duration between two events: The initialization event and the contact event. The initialization event occurs when the CoM crosses from inside to outside the support polygon, marking the start of falling. The contact event is defined as the moment when the body-ground contact force reaches its peak, indicating impact. During the 5-second trial, these events determine the model’s balance status: the initialization event signals the start of a fall, while the contact event marks ground collision. Beyond healthy conditions, we model impaired balance resulting from muscle injury by simulating reduced muscular function. Specifically, we restrict the force-generating capacity of the left rectus femoris (RF) muscle to replicate the effects of injury, as reported in \cite{unuvar2023comparison}.


\subsection{Exoskeleton-Assisted Standing}
\label{sec:exoskeleton}

\begin{wrapfigure}{r}{0.35\textwidth}
  \begin{center}
    \includegraphics[width=0.35\textwidth]{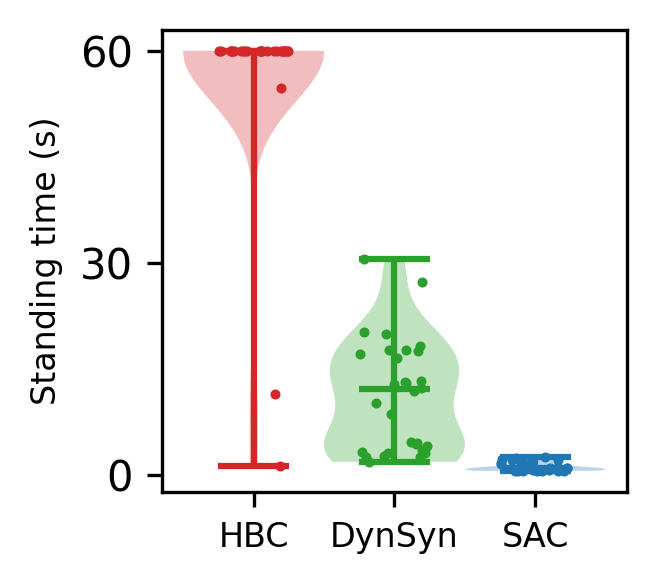}
  \end{center}
  \caption{Standing durations achieved by three algorithms. 
  The evaluations for balancing were terminated at 60 seconds.}
  \label{fig:baseline}
  \vspace{-40pt}
\end{wrapfigure}

As illustrated in Figure~\ref{fig:standing}, the high-level planner in \algo~ concurrently optimizes control strategies for balance-assistive devices such as a hip exoskeleton, enabling efficient simulation and design of exoskeletal systems. Assisted balance is simulated by applying torque actuation at hip joints—a widely accepted approach for modeling exoskeleton effects via externally applied forces \cite{dembia2017simulating}. To enhance assisted balance performance, we optimize the exoskeleton design parameters $p$ (e.g., control gains). Formulated as a black-box optimization problem, we apply Bayesian optimization to improve the exoskeleton’s posture correction policy.






\section{Experiments}
\label{sec:experiments}

In the experiments, we first evaluate the control performance of the \algo~algorithm for a high-dimensional whole-body standing task, with biomechanical fidelity check of the simulation results. We then generate balance and fall behaviors and conduct comprehensive analysis on healthy, injured, and exoskeleton-assisted conditions.


\subsection{Performance Evaluation of the \algo~Algorithm}

\begin{wrapfigure}{r}{0.55\textwidth}
  \begin{center}
    \includegraphics[width=0.55\textwidth]{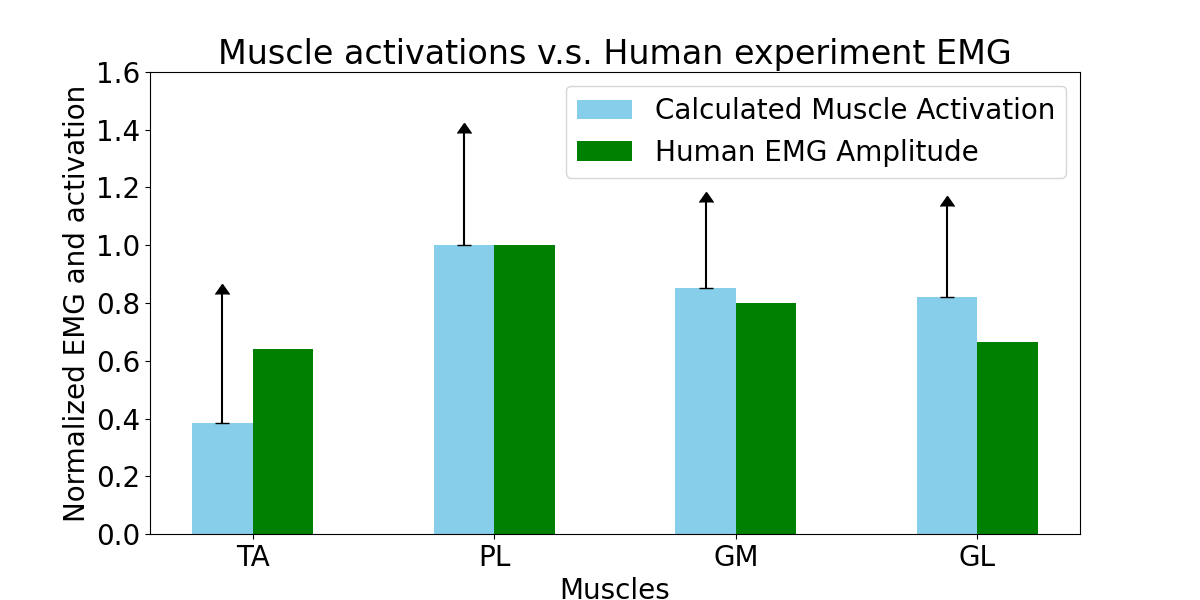}
  \end{center}
  \caption{Comparisons between calculated muscle activations and real human EMG. Simulated muscles activity (blue) of the standing task in normalized root mean square values. Experimental electromyography (EMG) signals of a human subject (green). Four experimentally measured muscles: TA, tibialis anterior; PL, peroneus longus; GM, gastrocnemius medialis; GL, gastrocnemius lateralis.}
  \label{fig:emgs}
\end{wrapfigure}

We compare \algo~against competitive DRL-based control baselines on the full-body balance control task. Our selected baselines include Dynamic Synergy Representation (DynSyn), the leading DRL-based control method for the control musculoskeletal systems \cite{he2024dynsyn}, and Soft Actor Critic (SAC) \cite{haarnoja2018softactorcriticoffpolicymaximum}. As shown in Figure~\ref{fig:baseline}, our approach successfully achieves static balance control on the full-body musculoskeletal system, significantly outperforming both baseline methods and attaining the maximum standing duration in most trials. We carried out a fidelity check on our simulated control results by comparing the muscle activation calculated by \algo~ with the human experimental results reported in  \cite{blaszczyszyn2019semg, yamanaka2023emg}. Figure \ref{fig:emgs} shows that the simulation results demonstrate consistency with real-world experiments.


\begin{figure}[b]
    \begin{subfigure}[b]{0.25\linewidth}
      \centering
      \raisebox{1.4em}{\includegraphics[width=\linewidth]{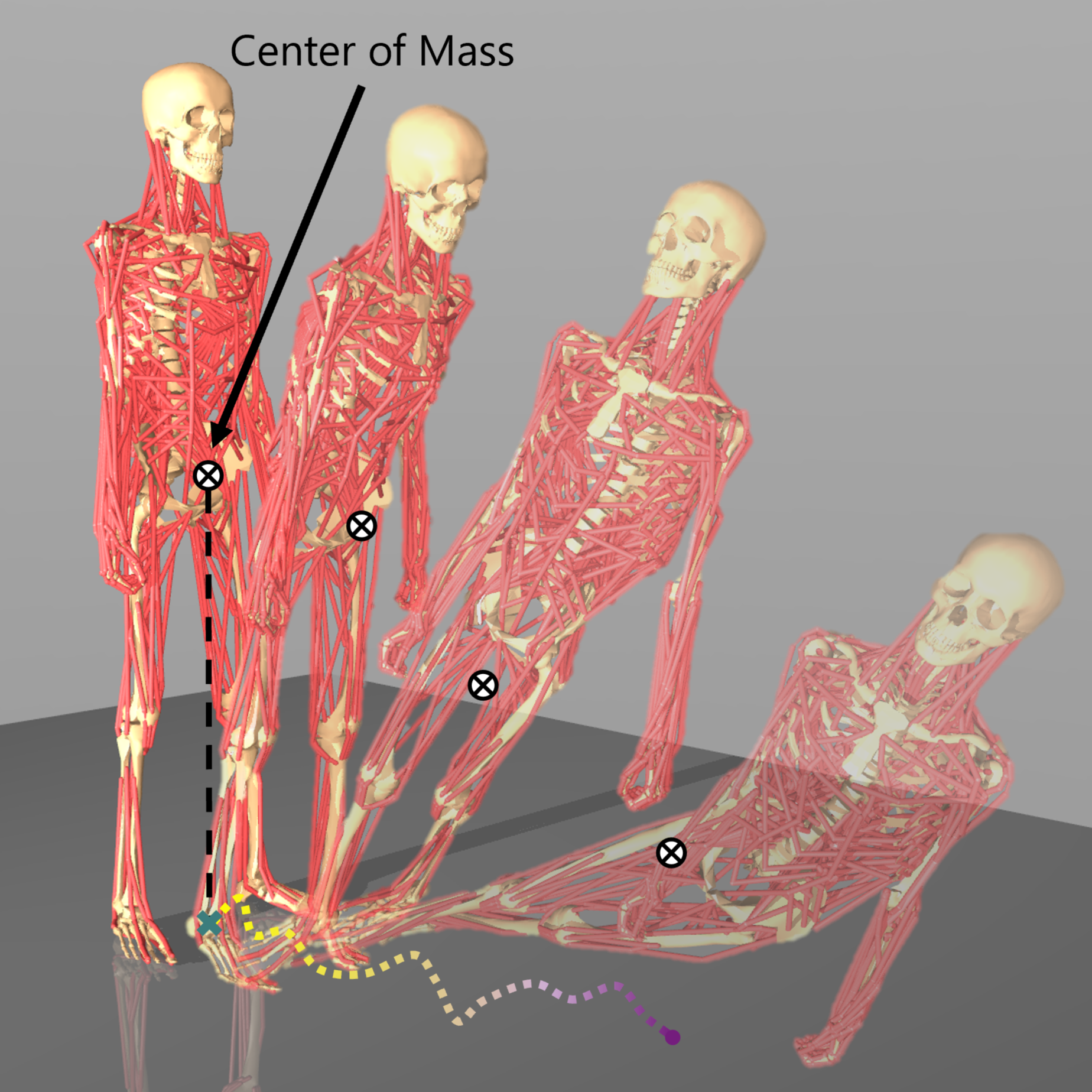}}
      \caption{}\label{fig:com_demo}
    \end{subfigure}
    \begin{subfigure}[b]{0.36\linewidth}
      \centering
      \includegraphics[width=\linewidth]{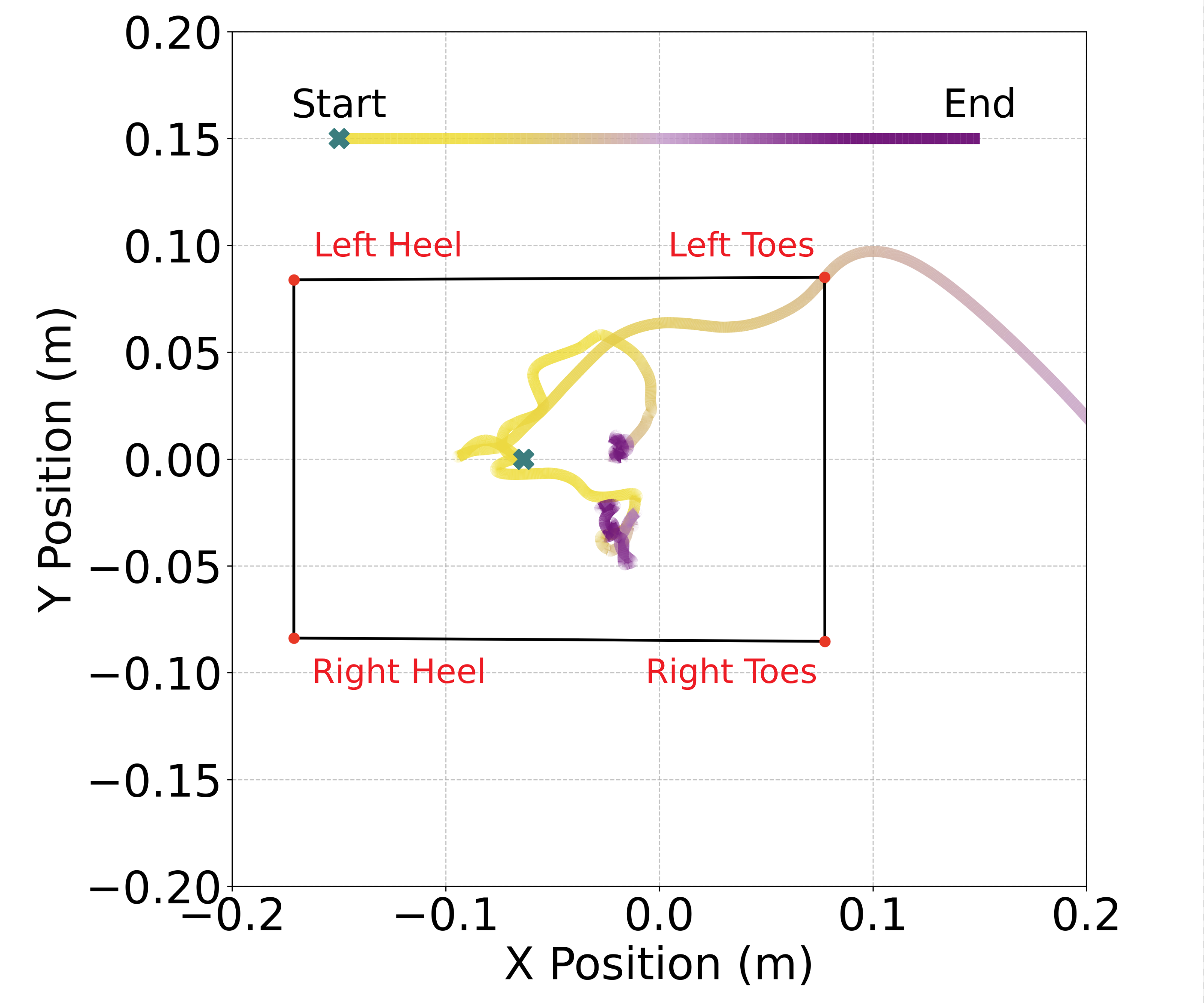}
      \caption{}\label{fig:com_trajectories}
    \end{subfigure}
    \hspace{-0.5em}  
    \begin{subfigure}[b]{0.37\linewidth}
      \centering
      \raisebox{0.4em}
      {\includegraphics[width=\linewidth]{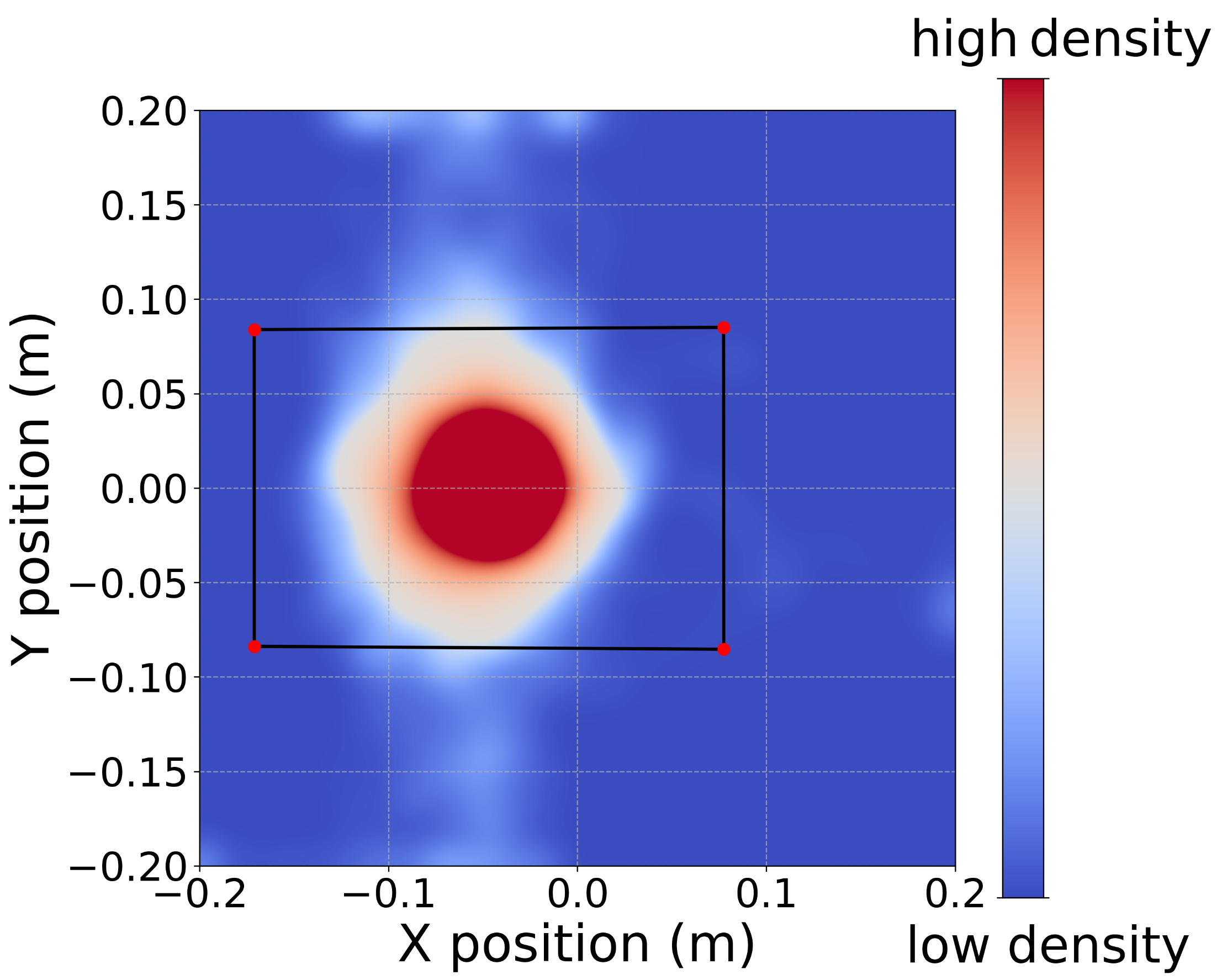}}
      \caption{}\label{fig:balanceregion_healthy}
    \end{subfigure}
  \caption{Human static balancing behavior dynamics. (a) Visualization of a human fall with the center of CoM projection trajectory. (b) CoM trajectories during balance and falling, with each trajectory color-coded over time from yellow to purple.  (c) Density plot of CoM coordinates during successful balance trials. The red area highlights the balance region, representing the highest density of CoM positions associated with stable balancing dynamics.}
  \label{fig:bothheatmap}
  
\end{figure}

\subsection{Study of Balance Dynamics}
\label{sec:comstudy}

In this setion, we study the dynamical behavior during balance control over a healthy model. A total of 2,800 falling trajectories were collected, where similar data collection in real world will be very difficult to achieve.
For all plots on the X-Y coordinate plane, the initial support polygon is demonstrated by connecting the starting positions of heels and toes. 

\textbf{CoM trajectories during balancing.} 
As discussed in Section \ref{sec:relatedwork}, CoM position and movement are important metrics to investigate bipedal balance.  We observe horizontal CoM position to track the subtle, dynamic movements of human body during balance as demonstrated in Figure \ref{fig:com_demo}. Representational CoM trajectories are visualized in Figure \ref{fig:com_trajectories}, where two of them are from successful balancing trajectories with the end converging near the starting point, and one from a falling trajectory. The CoM trajectories of balancing are usually irregular at the beginning, and converge to a smaller area, showing that balancing of a high-dimensional, muscle -controlled human body is dynamical and influenced by instant control and responses.

\textbf{Balance region achieved during stable standing.}
Although CoM series data are very noisy and diverse due to the dynamic nature of balance, it is possible to draw statistical insights from the overall distribution of CoM positions arrived during balance. 
We visualized the spatial density of CoM coordinates in 640,000 frames of successful trials during standing in Figure \ref{fig:balanceregion_healthy}. Similar data collection in real world will be very difficult to achieve.

\begin{wrapfigure}{r}{0.5\textwidth}
  \begin{center}
    \includegraphics[width=0.5\textwidth]{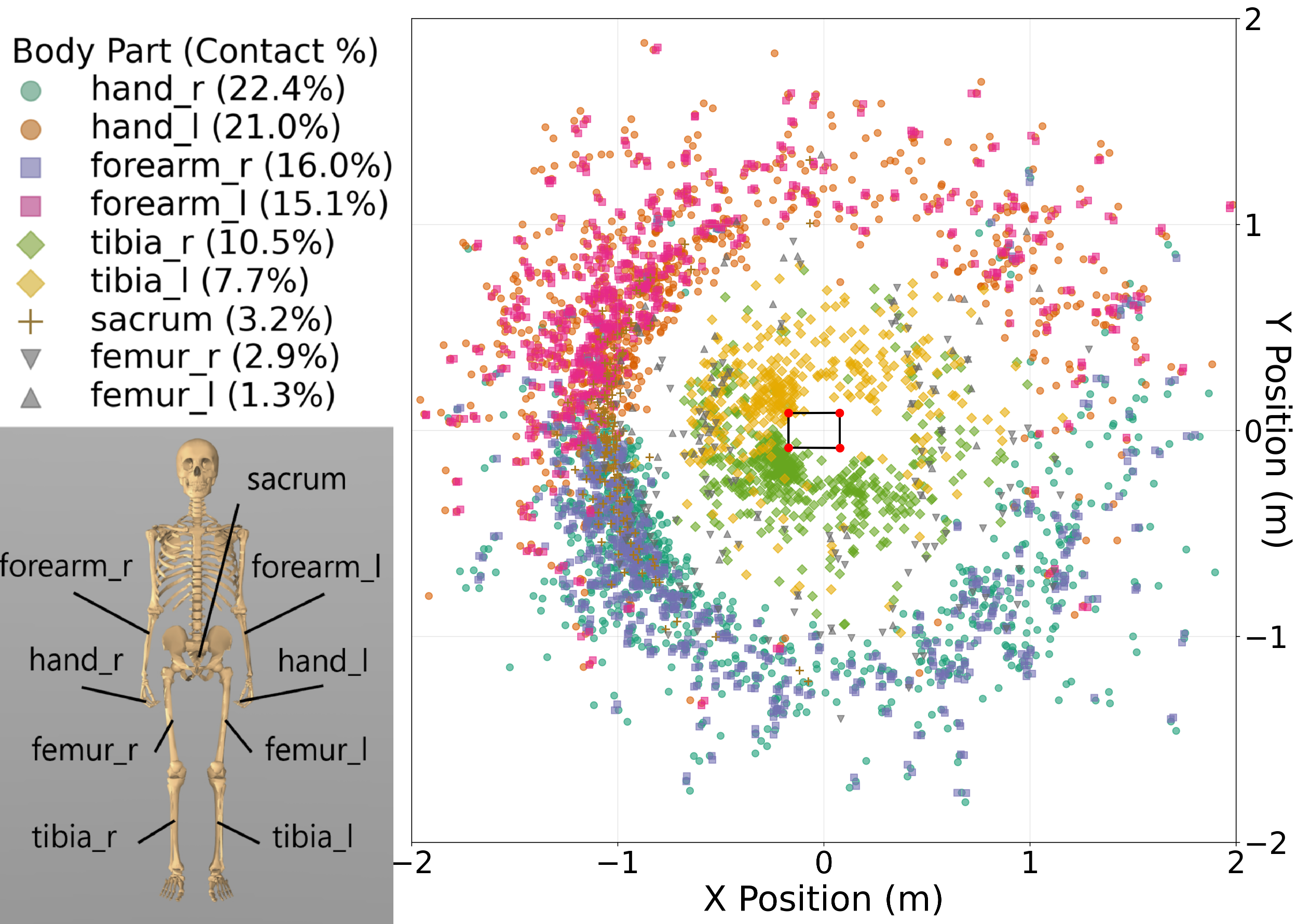}
  \end{center}
  \caption{
  Distribution of 2,895 collision positions in falling. We plot the body parts and their corresponding collision positions when the contact forces reach maximum value.
  }
  \label{fig:contact}
\end{wrapfigure}

\textbf{The distribution of collision positions following falls.} 
We recorded a total of 2,895 falls during the experiments and present statistical analyses based on the resulting falling behaviors.
We present statistical results from the collected falling behavior. Post-fall collision locations are closely associated with injury risk and offer insights that are difficult to obtain in real-world experimentation. As shown in Figure~\ref{fig:contact}, the most frequent collision regions during falls include the upper extremities (wrist and forearm), lower leg, and pelvis. This findings align with frequently injured sites in clinical data \cite{do2015fall, degoede2003fall}. Notably, 37\% of fall-related injuries involve broken or fractured bones \cite{do2015fall}, highlighting the biomechanical vulnerability of these impact sites. Our simulation framework facilitates detailed analysis of such injury-prone collision patterns, providing a biomechanically accurate testbed for the development of fall prevention and mitigation strategies.



\subsection{Study of Injury Effect on Balance}

\begin{figure}[bht]
    \begin{subfigure}[b]{0.17\linewidth}
      \centering
      {\includegraphics[width=\linewidth]{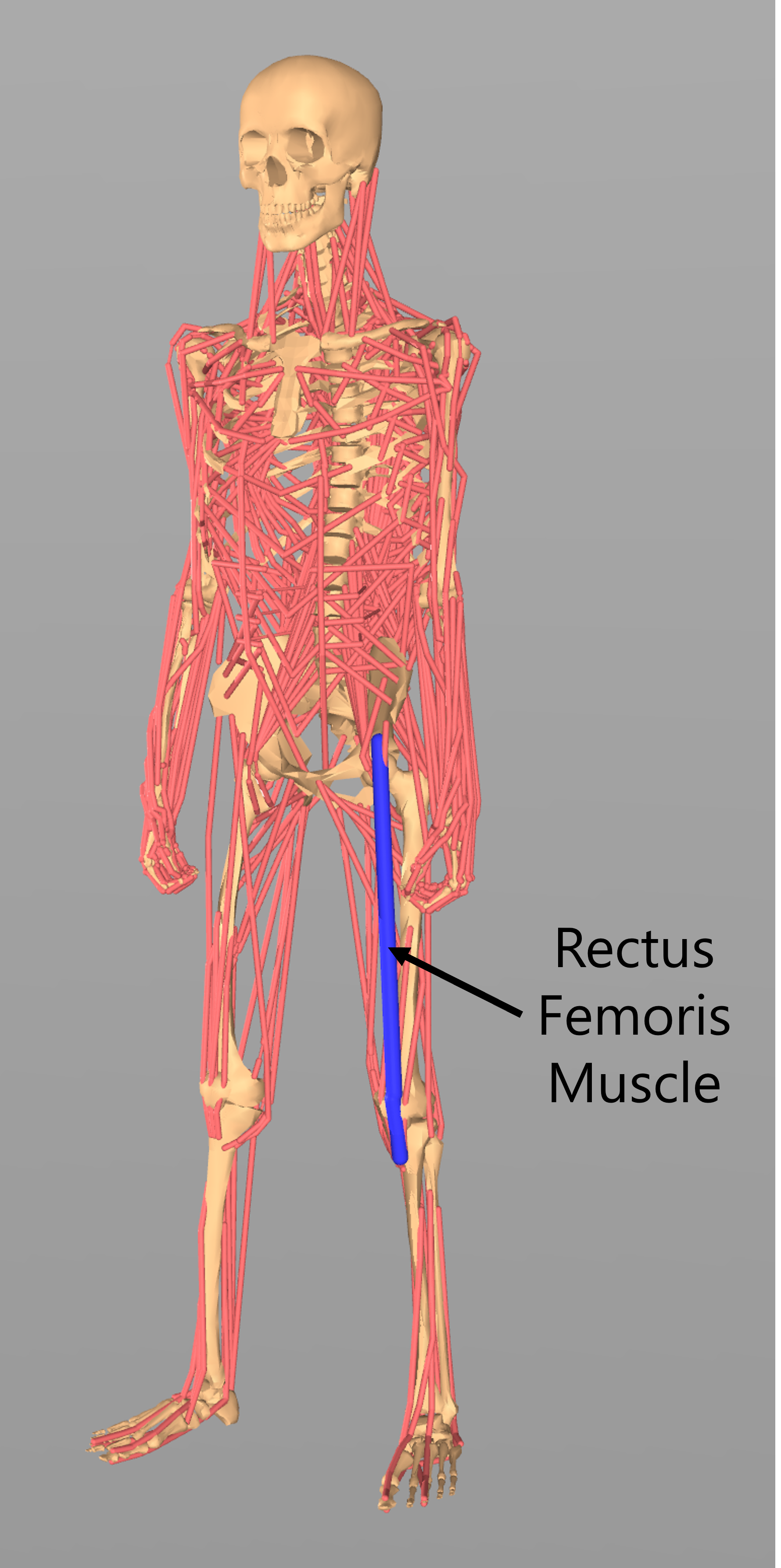}}
      \caption{}\label{fig:rf_demo}
    \end{subfigure}
    \hspace{-0.5em}  
    \begin{subfigure}[b]{0.42\linewidth}
      \centering
      \includegraphics[width=\linewidth]{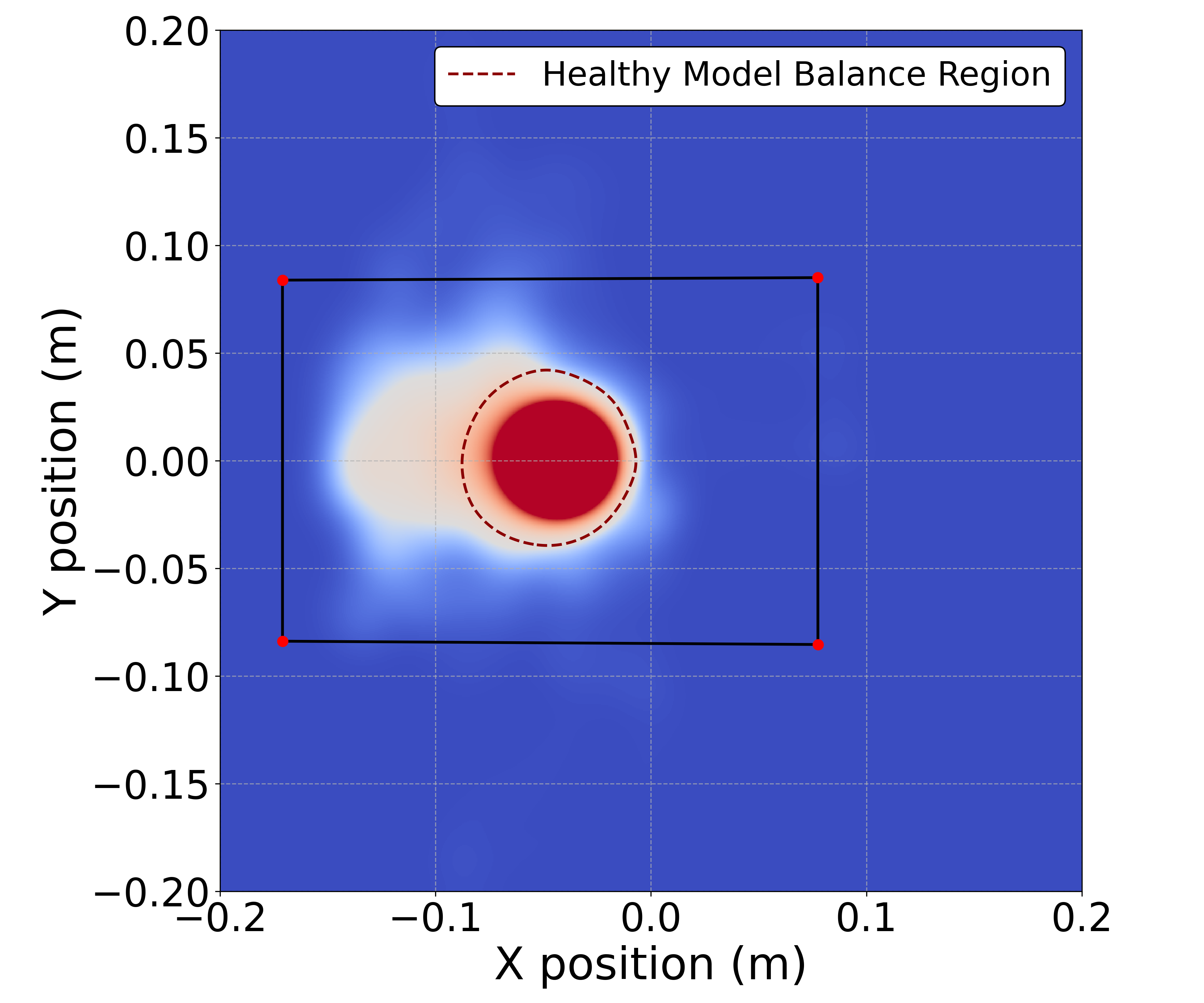}
      \caption{}\label{fig:balanceregion_injury}
    \end{subfigure}
    \hspace{-1.5em}  
    \begin{subfigure}[b]{0.42\linewidth}
      \centering
      \includegraphics[width=\linewidth]{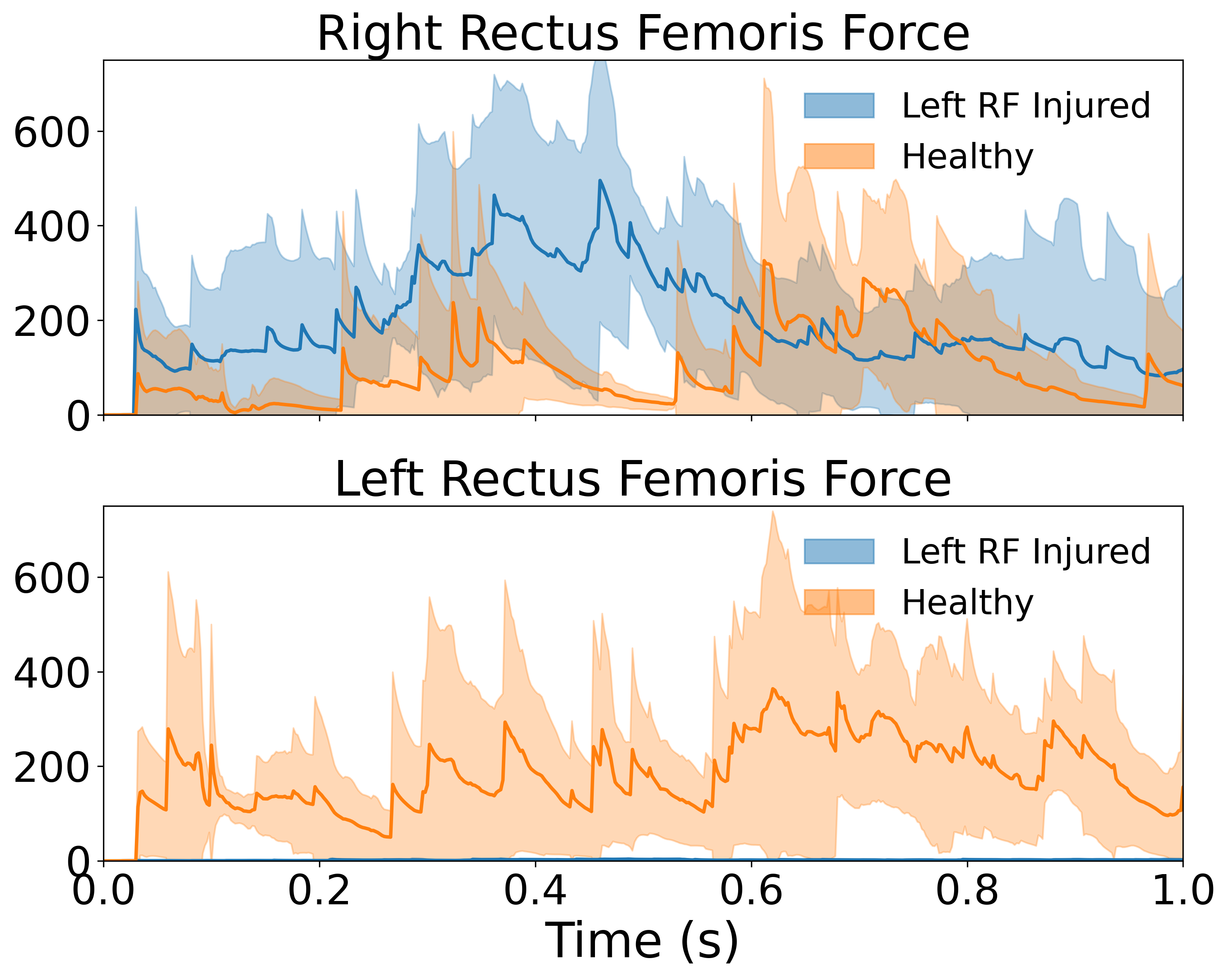}
      \caption{}\label{fig:rf_force}
    \end{subfigure}
  \caption{Balancing behavior under muscle injury conditions. (a) Illustration of the injured muscle (left rectus femoris).  (b) Density plot of CoM coordinates under muscle injury. The red dashed contour represents the balance region of the healthy model, highlighting the reduction in balance region due to impaired muscle function. (c) Muscle forces during standing; increased force in the right rectus femoris indicates compensatory activation in response to left side injury.}
  \label{fig:injury}

\end{figure}

With the injured model as introduced in Section \ref{sec:injury}, a total of 1280 falling trajectories were collected. We investigated the CoM behavior and muscle force responses under the injury model. Figure \ref{fig:balanceregion_injury} shows that the balance region shrinks with unilateral RF injury. The comparative analysis of the density distributions reveals two notable phenomena. First, the balance region exhibits a more concentrated pattern in the injured model, suggesting enhanced predictability and more conservative balancing strategy under injury. Second, there's an asymmetric expansion of the low-density region (white area) in the posterior direction, indicating an increased tendency of forward-leaning postural adjustments.  Figure \ref{fig:rf_force} shows a significant increase in right RF muscle force during balance maintenance, suggesting a weight shift to the right leg for stability.

\subsection{Simulation of Assisted Balance}

\begin{figure}[ht]
      \centering
      \begin{subfigure}[b]{0.23\linewidth}
      \centering
      \raisebox{0.6em}
      {\includegraphics[width=\linewidth]{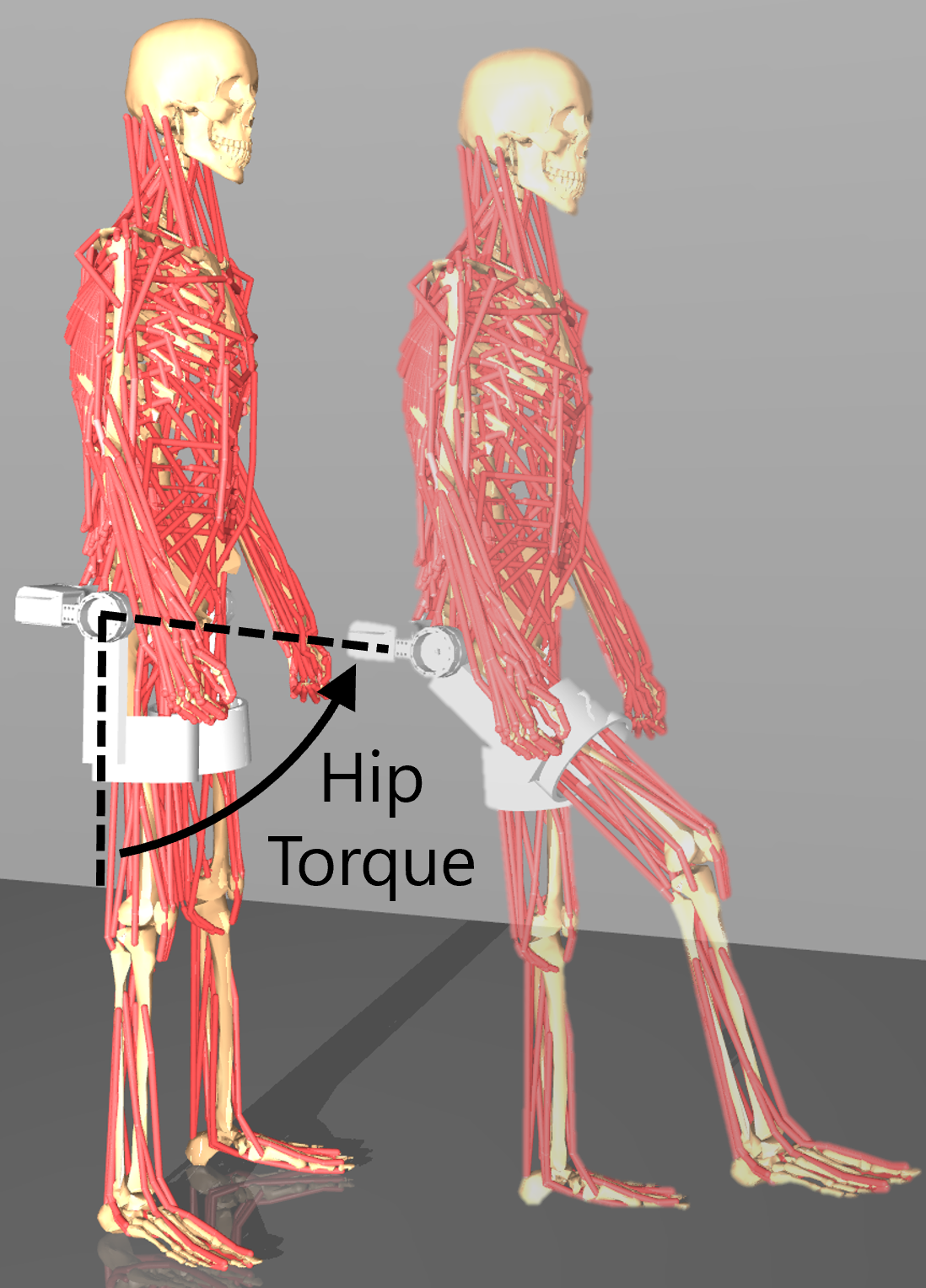}}
      \caption{}\label{fig:exo_illustrate}
    \end{subfigure}
      \begin{subfigure}[b]{0.35\linewidth}
      \centering
      \includegraphics[width=\linewidth]{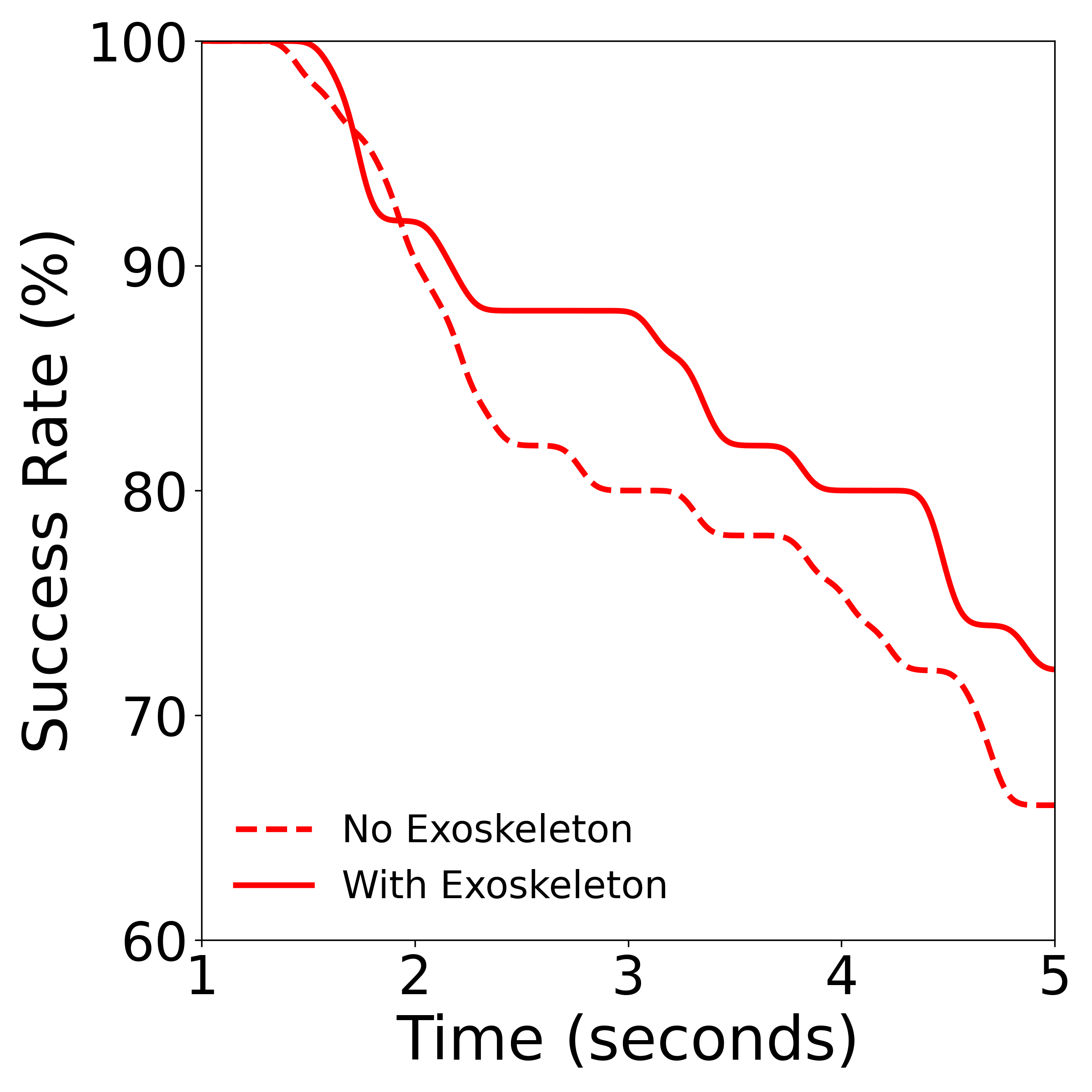}
      \caption{}\label{fig:success_rate}
    \end{subfigure}
    \begin{subfigure}[b]{0.39\linewidth}
      \centering
       \raisebox{0.6em}
      {\includegraphics[width=\linewidth]{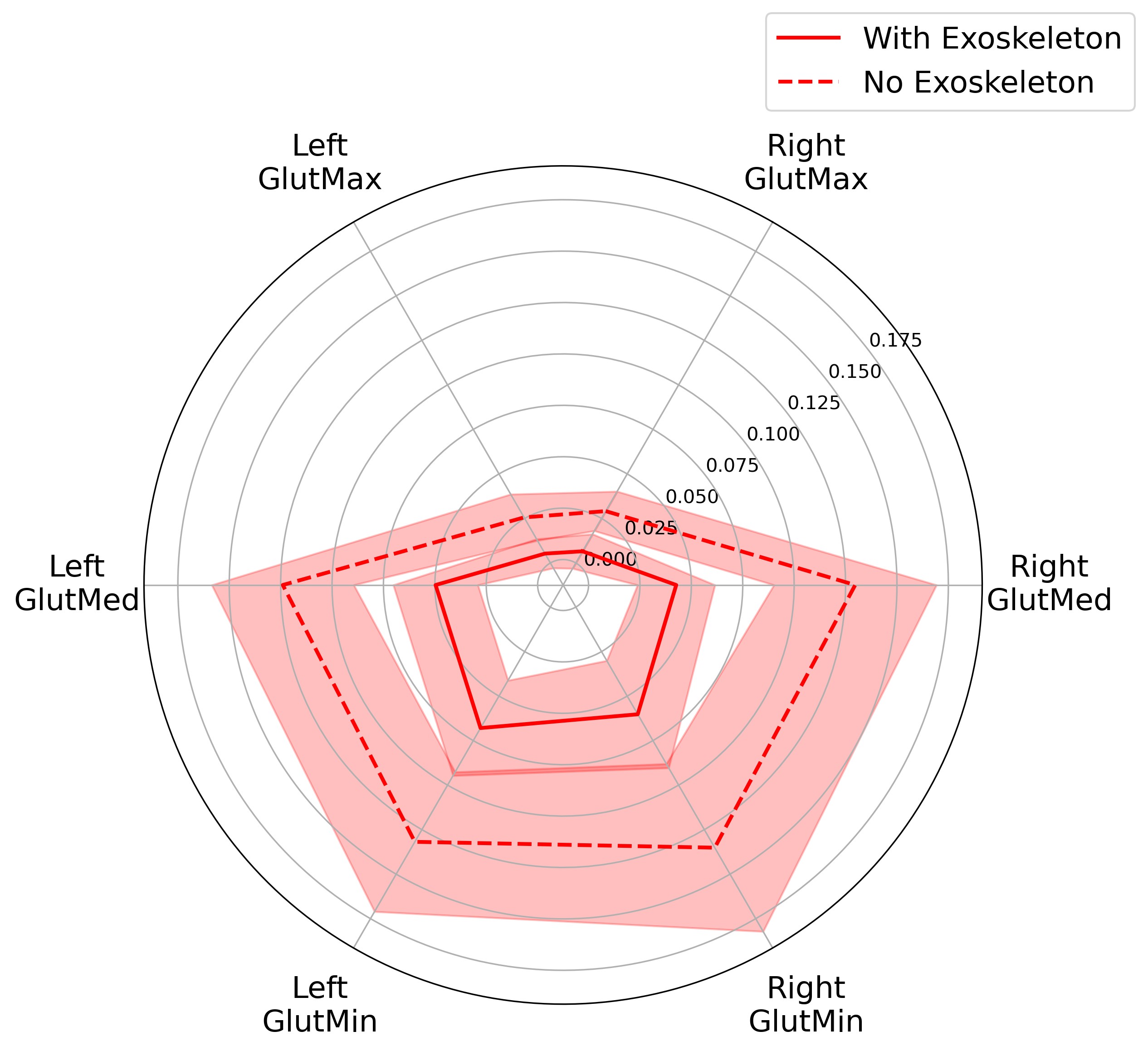}}
      \caption{}\label{fig:exo_radar}
    \end{subfigure}
  \caption{Balancing behavior under exoskeleton assistance. (a) Visualization of the hip exoskeleton device, using joint torques as actuators. (b) Success rate of maintaining standing posture, where the exoskeleton assistance helps better static balance. (c) Activation levels of gluteal muscle activations (GlutMax: gluteus maximus, GlutMed: gluteus medius, GlutMin: gluteus minimus) during the standing simulation, where muscle activation levels are reduced with exoskeleton assistance.}
  \label{fig:baselines}

\end{figure}

We simulated a assisted balance scenario with a hip exoskeleton as shown in Figure \ref{fig:exo_illustrate}.
We employed optimized exoskeleton parameters and validated the effectiveness of this exoskeleton control policy in a perturbation test: Models with and without exoskeleton assistance are pushed in random directions for 3 times with intervals of 1 second. As in Figure \ref{fig:success_rate}, exoskeleton-assisted balance achieved a higher success rate in maintaining balance under perturbation in the 5-second simulation.

In Figure \ref{fig:exo_radar}, we carried out a ablation study over the assisted balance by recording the muscle activations of the gluteus maximus, gluteus medius, and gluteus minimus, which play critical roles in lower-limb movement and postural stability.
We observe that muscle activation levels reduce with exoskeleton assistance, showing assistive devices' potential to save muscular effort and reduce metabolic cost. Such muscle-level data are difficult to obtain in traditional experiments due to the limitations of surface EMG and the inaccessibility of deep muscles. Our simulation pipeline enables validation of exoskeleton effects on balance maintenance. These findings suggest the utility of musculoskeletal simulations in evaluating and optimizing assistive device performance prior to costly physical prototyping and human subject testing.

\section{Conclusion}
\label{sec:conclusion}

We present a training-free hierarchical control method for studying full-body human bipedal balance using musculoskeletal simulation. Our Hierarchical Balance Control (\algo) approach enables effective planning and control by reducing the parameter space from muscle to joint level, and outperforms leading DRL-based methods. 

This work contributes to the understanding of human balance control by providing access to muscle-level insights that are difficult to obtain experimentally. Future work could explore more complex scenarios and develop targeted interventions for populations with balance limitations. While simulation has inherent limitations compared to real-world studies, our approach offers a complementary tool for investigating questions in human movement science that face practical experimental constraints.

\newpage
\section{Limitation}
\label{limitation}

This work demonstrates the challenge of maintaining human-like balance on a high-dimensional whole-body musculoskeletal model and proposes a hierarchical control method to achieve stable balance. While we achieves training-free balance control carried out comprehensive analysis under various conditions, several limitations remain.

First, although \algo~outperforms DRL baselines, we used an intuitive cost function design without extensive tuning. In contrast to DRL methods that often rely on carefully engineered reward functions, our approach might benefit from more sophisticated cost function designs to further improve performance and biological fidelity. Second, our simulations primarily focus on a standard healthy adult model. We did not extensively simulate vulnerable populations such as elderly individuals or those with specific neurological conditions that affect balance. These populations exhibit distinct biomechanical and neuromuscular characteristics that may lead to different balance control strategies and falling patterns, limiting the generalizability of our conclusions to these important user groups. Finally, our exoskeleton simulation uses simplified torque actuation rather than modeling the complete mechanical interface between the device and human tissue. This approximation, while computationally efficient, may not fully capture the complex effects that occur in real human-exoskeleton systems.


\section{Acknowledgment}
This work is funded by STI2030-Major Projects-2022ZD0209400, NSFC-62206152, and Tsinghua Dushi Fund.

\newpage
\bibliography{root}

\section{Supplementary material}

\subsection{Implementation detail}

\textbf{MJPC environment.} We used MuJoCo MPC \cite{howell2022mjpc} in our experiments to deploy \algo~in simulation time. With the vey high dimensionality of the MS-Human-700 model, the simulation speed was set to 10\%. A 5-second experiment will cost 50 seconds, which is much shorter than the training time of reinforcement learning methods. \algo~experiments used 16-core CPU + RTX 3080 GPU (600ms/32-step, 16-particle planning). 

\textbf{DRL experiments set up.} RL training used DynSyn \cite{he2024dynsyn} on 224-core CPU + RTX 4070 GPU (CPU simulation was the bottleneck; adding more GPUs doesn’t accelerate training). RL polices trained for $>$1 day still fails at stable standing, while \algo~can generate 2000 trajectories within 1 day, enabling significant speedup.

\subsection{Intuitive cost function design}

Standing balance could be described by the dynamics of CoM and ZMP. We set the control goal of human standing simulated by a musculoskeletal model as follow: (1) the horizontal velocity of the CoM should be near 0; (2) the ZMP should fall inside the support region, represented by a posture that has no trend of flipping over. The model may take a step to maintain balance, but the initial position of the model is kept the same for all experiments.

Considering these basic mechanical requirements of balancing, and hints of a desirable standing posture, we arrange components of the cost function $C$ as follows. The weights were tuned to $(w_H, w_R, w_{P_c}, w_{V_c}, w_I) = (300, 300, 300, 10, 1)$.
\begin{equation}
    C = w_H \cdot C_H + w_R \cdot C_R + w_{P_c} \cdot C_{P_c} + w_{V_c} \cdot C_{V_c} + w_I \cdot C_I
\end{equation}

(1) \textbf{Height, $C_H$}: This term penalizes the difference between body height and the initial height, which represents the status of the upright, not-flipping-over posture. It reaches 0 when the difference between head and feet of the musculoskeletal model is the same as its initial value. The height cost represents the status of the upright, not-flipping-over posture. $H_{head}$ is the height of the head, and $H_{feet}$ is the average height of left and right feet.
\begin{equation}
    C_H = z_{head} - z_{feet}
\end{equation}

(2) \textbf{Rotation, $C_R$}: This term penalizes large rotational angle of the model's upper body. $\vec{t}$ represents the unit vector pointing from the pelvis to the head, and $\vec{z}$ is the vertical unit vector.
 \begin{equation}
    C_R= 1 - \vec{t} \cdot \vec{z}
\end{equation}

(3) \textbf{CoM position $C_{P_c}$}: This term penalizes the mismatch between the horizontal CoM and the center of the support region, which is the convex hull formed by the heels and toes. $(x_{CoM}, Y_{CoM})$ refers to the horizontal position of the center of mass, while $(x_{feet}, y_{feet})$ is the centroid of the polygon formed by the left and right heels and toes. 
\begin{equation}
    C_{P_c} = |(x_{CoM}, y_{CoM}) - (x_{feet}, y_{feet})|
\end{equation}

(4) \textbf{CoM velocity, $C_{V_c}$}: This term penalizes large horizontal velocity of CoM. 
\begin{equation}
    C_{V_c} = |(\dot{x}_{CoM}, \dot{y}_{CoM})|
\end{equation}

(5) \textbf{Imitation, $C_I$}: This term penalizes the mismatch between the joint positions and joint angles of a desirable, natural standing posture. $q$ is the joint angles of the model, and $q_{ref}$ is the full-body joint angles of a natural standing posture.
\begin{equation}
    C_I = |q-q_{ref}|
\end{equation}

\subsection{Exoskeleton control policy and optimization} 

We implemented a weighted postural PD control over the joint torque actuation placed at the left and right hip flexion joints. The high-level planner (as discussed in Section \ref{sec:hbc}) is adapted to plan an extra target posture indicating the overall leaning direction of the body, represented by the tilt angle of the pelvis. The control policy of the exoskeleton torque actuation is separated into two parts mixed by a weight: (1) Hip flexion joint angle PD control. (2) Postural PD control. The control is formulated as follow:

\begin{equation}
        \tau_e^i = (1 - w) \cdot (k_{p_e}\cdot(q^{*}_i-q_i) + k_{d_e}\cdot(0 - \Dot{q}_i)) + w \cdot (k_{p_{t}} \cdot (q^{*}_{t}-q_{t}) + k_{d_{t}}\cdot(0 - \Dot{q}_{t}).
        \label{equ:exo}
    \end{equation}

$i = 1, 2$ represents the left or right side respectively. $\tau_e^i$ is the torque actuation value. $k_{p_e}$ and $k_{p_e}$ are the joint angle PD control constants, while $k_{p_t}$ and $k_{p_t}$ are the postural PD control constants over the tilt angle of the pelvis. $q^{*}_i$ and $q^{*}_t$ are the target values of the hip joint angles and the pelvis tilt angle. $w$ is the weight between the two PD control policies.

\begin{figure*}[h]
  \centering
  \includegraphics[width=.6\linewidth]{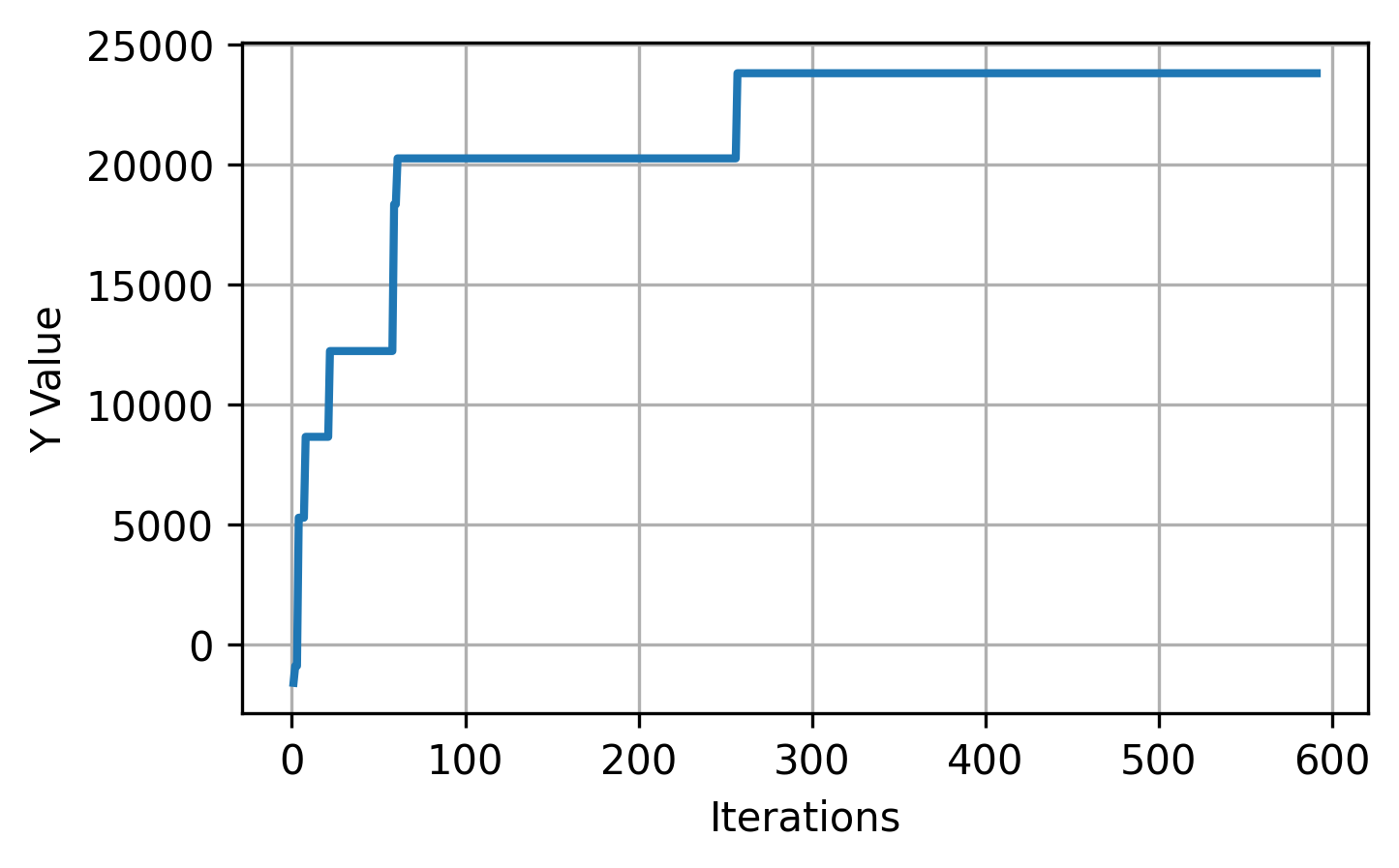}
  \caption{Exoskeleton parameter optimization results. The curve shows the best objective value during optimization.}
  \label{fig:bo}
\end{figure*}

We found that the assistive effect of the exoskeleton was very sensitive to the $k_{p_e}$, $k_{p_e}$, $k_{p_t}$ and $w$ values. Therefore, we carried out Bayesian optimization (BO) to determine a set of parameter to ensure performance across trials. We define $\boldsymbol{x} = (k_{p_e}, k_{p_e}, k_{p_t}, w)$, and formulate the parameter search as a black-box optimization problem:
\begin{align}
    \max_{\boldsymbol{x}\in \mathcal{X}} f(\boldsymbol{x})=\mathbb{E}_{\boldsymbol{x}} \left[-\sum_{t=0}^{T-1} C(s_t, u_t)\right]
\end{align}
where the objective $f(\boldsymbol{x})$ is the negative cumulative cost function under parameter $\boldsymbol{x}$, averaged over 5 independent trials. We assume the observation noise is i.i.d Gaussian: $y = f(\boldsymbol{x}) + \eta, \eta\sim \mathcal{N}(0, \sigma^2)$.
Given sampled data $\mathcal{D} = \{(\boldsymbol{x}_1, y_1), \cdots, (\boldsymbol{x}_n, y_n)\}$
We use Gaussian process to model the objective function with posterior mean and covariance estimation under kernel function $k$:
\begin{equation}
\begin{split}
    \mu_n(\boldsymbol{x})& = \boldsymbol{k}_n(\boldsymbol{x})^T(\boldsymbol{K}_n+\sigma^2\boldsymbol{I})^{-1}\boldsymbol{y}_n \\
    k_n(\boldsymbol{x},\boldsymbol{x}')& = k(\boldsymbol{x},\boldsymbol{x}')-\boldsymbol{k}_n(\boldsymbol{x})^T (\boldsymbol{K}_n+\sigma^2\boldsymbol{I})^{-1} \textbf{k}_n(\boldsymbol{x}')\\
    \sigma^2_n(\boldsymbol{x}) &= k_n(\boldsymbol{x},\boldsymbol{x}),
\end{split}
\label{eq:gp}
\end{equation}
where $\boldsymbol{k}_n(\boldsymbol{x}) =[k(\boldsymbol{x}_1,\boldsymbol{x}), ..., k(\boldsymbol{x}_n,\boldsymbol{x})]$ is the covariance between $\boldsymbol{x}$ and sampled points, $\boldsymbol{K}_n$ is the covariance of sampled positions:  $[k(\boldsymbol{x},\boldsymbol{x}')]_{\boldsymbol{x},\boldsymbol{x}' \in X_n}$. Given the GP posterior, we optimize the Expected Improvement acquisition function to sample the next parameter \cite{jones1998efficient}:
\begin{align}
    \text{EI}_n(\boldsymbol{x}) & = \mathbb{E}[[f(\boldsymbol{x})-f^*_n, 0]^+]\\
    & =(\mu_{n}(\boldsymbol{x})-f^*_n)\Phi(\frac{\mu_n(\boldsymbol{x})-f^*_n}{\sigma_t(\boldsymbol{x})}) + \sigma_n(\boldsymbol{x})\Phi(\frac{\mu_n(\boldsymbol{x})-f^*_n}{\sigma_n(\boldsymbol{x})}),
\end{align}
where $f^*_n = \max_{i\in n} y_i$. Our sequentual optimization procedure is illustrated in Algorithm \ref{alg: BO}. We implemented the overall BO procedure based on BoTorch \cite{balandat2020botorch}. Figure \ref{fig:bo} shows the optimization performance over 600 iterations. The effectiveness of this exoskeleton control policy is validated in a perturbation test: Models with and without the exoskeleton assistance are pushed in random directions for 3 times with intervals of 1 second. We observe that under the optimize parameter, the model with the assistance achieves a higher success rate in maintaining balance under perturbation through the 5-second simulation.

\begin{algorithm2e}[t]
\SetAlgoLined
\DontPrintSemicolon
\LinesNumbered
\caption{Exoskeletal parameter search via Bayeisan optimization}
    \label{alg: BO}
        \KwIn{ Initial dataset $\mathcal{D}_0$, Gaussian process $\mathcal{M}_0$}
        \For{$n = 1, 2, \cdots$}{
            Update $\mathcal{M}_t$ based on $\mathcal{D}_{n-1}$\;
            $\boldsymbol{x}_n \leftarrow \text{argmax}_{\boldsymbol{x} \in \mathcal{X}}\text{EI}_n(\boldsymbol{x}))$\;
            $y_n \leftarrow f(\boldsymbol{x}_n) + \eta$\;
            $\mathcal{D}_{n} \leftarrow \mathcal{D}_{n-1}\cup(\boldsymbol{x}_{n}, y_{n})$\;
            }
\end{algorithm2e}

\subsection{Musculoskeletal Model Dynamics}

\textbf{Rigid-body dynamics} 

The musculoskeletal model used in this work is the MS-Human-700 model \cite{tsht}. It comprises of 90 rigid body segments, 206 joints and 700 muscle-tendon units. By actuating its 700 muscle-tendon units, the model can be controlled and perform human-like tasks. The dynamics of the model can be formulated as follow:
    \begin{equation}
        M(q)\Ddot{q} + c(q,\Dot{q}) = J_{m}^T f_m(act) + J_c^T f_c + \tau_{ext}.
    \label{dynamics}
    \end{equation}
On the left side of the equation, $q$ stands for generalized coordinates of joints, $M(q)$ stands for the mass distribution matrix, and $c(q,\Dot{q})$ stands for Coriolis and the gravitational force. On the right side, $J_{m}$ and $J_{c}$ stand for Jacobian matrices that map forces to the generalized coordinates, $f_{c}$ is the constraint force, $f_m(act)$ stands for actuator forces generated by muscle-tendon units determined by muscle activations ($act$), and $\tau_{ext}$ stands for all external torque when interacting with environments.

MS-Human-700 model is implemented in the MuJoCo physics engine \cite{mujoco}. The actuators of the model in this work are 700 Hill-type \cite{zajac1989muscle} muscles. The actuator force generated by each muscle-tendon unit, and the temporal relation between muscle activation $act$ and the input control signal of the musculoskeletal model $u$ can be decided by the following equations:
\begin{equation}
    f_m(act)=f_{max}\cdot [F_{l}(l_m)\cdot F_v(v_m)\cdot act + F_p(l_m)].
    \label{equ:flv}
\end{equation}
\begin{equation}
    \frac{\partial act}{\partial t} = \frac{u-act}{\tau(u,act)}, 
\label{equ:muscle_dynamic}
\end{equation}
In Eq. (\ref{equ:flv}), $F_l$ and $F_v$ represent force-length and force-velocity functions which are actuator gains, $F_p$ is the passive force that works as actuator bias, and $l_m, v_m$ are normalized length and normalized velocity of the muscle. $f_{max}$ is the maximum isometric muscle force as specified in the model. In the first-order nonlinear system described by Eq. (\ref{equ:muscle_dynamic}), muscle activation $act$ is calculated. The time parameter $\tau$ is computed following Millard et al. \cite{millard2013flexing}. $\tau$ is the time constant related to the latency in activation and deactivation. 

\textbf{Muscular Inverse Dynamics}

To tackle the challenge of musculoskeletal model postural control posed by its very high dimensionality and non-trivial nature, we try to avoid directly manipulating the control of 700 muscles, but approach a proper target pose defined by joint angles of the model. The length of muscles is one characteristic of the muscles that can be determined by a given set of joint angles on MS-Human-700 model. A PD control-like formulation over muscle lengths can be used to derive desired muscle forces. A normalized PD control is implemented as
    \begin{equation}
        f_m = min(0,k_p\cdot(l^{*}_m-l_m)/l_{range} + kd\cdot(0 - \Dot{l}_m)),
    \end{equation}

where $f_m$ stands for muscle forces, $k_p$ and $k_d$ are PD control gains, $l_m$ stands for actual muscle lengths, $\Dot{l}_m$ stands for muscle velocities and $l^{*}_m$ is target muscle lengths. The difference between the maximum muscle length and the minimum muscle length of each muscle, $l_{range}$, is used to normalize muscle lengths and stablize control effect.  

To fully control the muscle-actuated musculoskeletal model after obtaining desired muscle forces, inverse dynamics of the muscles are applied. 
We assume that a target muscle activation, determined by desired muscle forces, should be approached in the next timestep. 
Eq. (\ref{equ:flv}) and Eq. (\ref{equ:muscle_dynamic}) can therefore be reformulated as follows:
    \begin{equation}
        act^{*} = \frac{f^{*}/f_{max} - F_p}{F_l \cdot F_m},
    \end{equation}
    \begin{equation}
        act^{*} = act + ts \cdot \frac{u - act}{\tau},
    \label{equ:act}
    \end{equation}
where $act^{*}$ is the target muscle activation and $ts$ is the simulation timestep. According to MyoSuite \cite{MyoSuite2022}, the discrete time constant can be approximated, making it possible to obtain the desired input control in closed form $u = \mathcal{PD} (s, z^*)$, where $z^*$ denotes the desired target joint angles.

\end{document}